\newcommand{\xbD}{\Delta}

\newcommand{\xbS}{\Sigma}

\newcommand{\xba}{\alpha}
\newcommand{\xbb}{\beta}

\newcommand{\xbm}{\mu}

\newcommand{\xCf}{\hspace{0.1em}}

\newcommand{\xcc}{\subseteq}

\newcommand{\xco}{\vee}
\newcommand{\xcp}{\rightarrow}

\newcommand{\xcs}{\cap}
\newcommand{\xcu}{\wedge}

\newcommand{\xDH}{\item }

\newcommand{\xEI}{\begin{itemize}}
\newcommand{\xEJ}{\end{itemize}}

\newcommand{\xEh}{\begin{enumerate}}

\newcommand{\xEj}{\end{enumerate}}

\newcommand{\xEn}{\begin{description}}
\newcommand{\xEp}{\end{description}}

\newcommand{\xfI}{\mbox{I}}

\newcommand{\bl}{\begin{lemma} \rm}
\newcommand{\el}{\end{lemma}}
\newcommand{\br}{\begin{remark} \rm}
\newcommand{\er}{\end{remark}}
\newcommand{\be}{\begin{example} \rm}
\newcommand{\ee}{\end{example}}
\newcommand{\bco}{\begin{corollary} \rm}
\newcommand{\eco}{\end{corollary}}
\newcommand{\bc}{\begin{claim} \rm}
\newcommand{\ec}{\end{claim}}
\newcommand{\bfa}{\begin{fact} \rm}
\newcommand{\efa}{\end{fact}}
\newcommand{\bp}{\begin{proposition} \rm}
\newcommand{\ep}{\end{proposition}}
\newcommand{\bd}{\begin{definition} \rm}
\newcommand{\ed}{\end{definition}}
\newcommand{\bcs}{\begin{construction} \rm}
\newcommand{\ecs}{\end{construction}}
\newcommand{\bcd}{\begin{condition} \rm}
\newcommand{\ecd}{\end{condition}}
\newcommand{\bt}{\begin{theorem} \rm}
\newcommand{\et}{\end{theorem}}
\newcommand{\bn}{\begin{notation} \rm}
\newcommand{\en}{\end{notation}}
\newcommand{\bfi}{\begin{bild} \rm}
\newcommand{\efi}{\end{bild}}
\newcommand{\bsta}{\begin{statement} \rm}
\newcommand{\esta}{\end{statement}}
\newcommand{\bcom}{\begin{comment} \rm}
\newcommand{\ecom}{\end{comment}}
\newcommand{\bdia}{\begin{diagram} \rm}
\newcommand{\edia}{\end{diagram}}

\newcommand{\bfc}{\begin{figure}[htb] \begin{center}}
\newcommand{\efc}{\end{center} \end{figure}}

\documentclass{article}

\usepackage{amssymb,latexsym,epic,eepic,rotating}

\title{A pre-semantics for counterfactual conditionals and similar logics
}

\author{Karl Schlechta
\thanks{
schcsg@gmail.com - https://sites.google.com/site/schlechtakarl/ -
Koppeweg 24, D-97833 Frammersbach, Germany}
\thanks{
Retired, formerly: Aix-Marseille Universit\'{e}, CNRS, LIF UMR 7279, F-13000
Marseille, France
}
}

\begin{document}

\newtheorem{lemma}{Lemma}[section]
\newtheorem{theorem}[lemma]{Theorem}
\newtheorem{proposition}[lemma]{Proposition}
\newtheorem{corollary}[lemma]{Corollary}
\newtheorem{claim}[lemma]{Claim}
\newtheorem{fact}[lemma]{Fact}
\newtheorem{remark}[lemma]{Remark}
\newtheorem{definition}{Definition}[section]
\newtheorem{construction}{Construction}[section]
\newtheorem{condition}{Condition}[section]
\newtheorem{example}{Example}[section]
\newtheorem{notation}{Notation}[section]
\newtheorem{bild}{Figure}[section]
\newtheorem{comment}{Comment}[section]
\newtheorem{statement}{Statement}[section]
\newtheorem{diagram}{Diagram}[section]

\renewcommand{\labelenumi}
  {(\arabic{enumi})}
\renewcommand{\labelenumii}
  {(\arabic{enumi}.\arabic{enumii})}
\renewcommand{\labelenumiii}
  {(\arabic{enumi}.\arabic{enumii}.\arabic{enumiii})}
\renewcommand{\labelenumiv}
  {(\arabic{enumi}.\arabic{enumii}.\arabic{enumiii}.\arabic{enumiv})}

\maketitle

\setcounter{secnumdepth}{3}
\setcounter{tocdepth}{3}

\begin{abstract}

The elegant Stalnaker/Lewis semantics for counterfactual conditonals works
with distances between models. But human beings certainly have no tables
of models and distances in their head.

We begin here an investigation using a more realistic idea, based on
findings in neuroscience. We call it a pre-semantics, as its
meaning is not a description of the world, but of the corresponding structure in
the brain, which
is (partly) determined by the world it reasons about.

Our basic concept is that of a "picture" or "scenario" on the meaning level, a
group of connected neurons on the neural level.

\end{abstract}

\tableofcontents

%
%
%
\clearpage
\section{
Introduction
}
\subsection{
The Stalnaker/Lewis semantics for counterfactual conditionals
}

Stalnaker and Lewis, see e.g.
 \cite{Sta68},  \cite{Lew73},
gave a very elegant semantics to counterfactual conditionals, based on
minimal change.
To give meaning to the sentence ``if it were to rain, $ \xfI $ would take
an
umbrella'', we look at all situations (models) where it rains, and
which are minimally different from the present situation. If $ \xfI $ take
an
umbrella in all those situations, then the sentence is true.
E.g., situations where there is hurricane - and $ \xfI $ will therefore
not take an
umbrella - will, usually, be very different from the present situation.

This idea is very nice, but we do not think this way. First, we have no
catalogue of all possible worlds in out head. Thus, we will have to
compose the situations to consider from various fragments. Second,
classical reasoning, taken for granted in usual semantics, has an
``inference cost''. E.g., when reasoning about birds, we might know that
penguins are birds, but they might be too ``far fetched'', and forgotten.

It seems that human beings reason in pictures, scenes, perhaps prototypes,
but in relatively vague terms. We try to use a more plausible
model of this reasoning, based on neural systems, to explain
counterfactual conditionals. But the basic Stalnaker/Lewis idea is upheld.

First, we develop in the next section our ideas how we think. We will see
that
the main problems on our level are conceptual in nature.
It seems difficult to find elementary, atomic, pictures and corresponding
connected groups of neurons, so we postulate that there are none. We can
always analyse, look inside, and, conversely, combine. The right level
of abstraction depends on the
structure examined. In addition, there is no surface as opposed to an
inside,
as cells have, all inter-group connections seem to go from inside to
inside.

At the end, we come back to our initial problem, the meaning of
counterfactual conditionals.

Our ideas are very rudimentary, all details are left open. Still, we think
that it is a reasonable start.
We might be overly flexible in our concepts, but it is probably easier
to become more rigid later, than inversely.

(Note that the case of update is somewhat different. In update, the
question is how the actual world changes, and not how we believe the
world (and we) behave, as is the case here.)
\section{
Background
}
\subsection{
Introduction
}

We discuss here our ideas how human beings think: not in propositions and
logical operators, or in
models, but in pictures, scenarios, prototypes, etc. On the neural level,
such pictures correspond to groups of neurons.

We will be vague, on the meaning level, as well as the neural
level. This vagueness results in flexibility,
the price to pay are conceptual difficulties.

We have three types of objects:
 \xEh
 \xDH pictures or groups (of neurons),
 \xDH connections,
 \xDH attention.
 \xEj

Groups will be connected areas of the brain, corresponding to some
picture,
groups may connect to other groups with different types of connectors,
which
may be positive or negative. Finally, attention
focusses on groups or parts of groups, and their connections, or part
thereof.

Groups do not necessarily correspond to nodes of graphs, as they are not
atomic, and they can combine to new groups. Attention may hide
contradictions,
so the whole picture may contain contradictions.

We use the word ``group'' to designate
 \xEI
 \xDH
on the physiological level a (perhaps only momentarily) somehow connected
area of the brain, they may be formed and dissolved dynamically,
 \xDH
on the meaning level a picture, scene (in the sense of conscious scene),
a prototype (without all the connotations the word ``prototype'' might
have),
any fragment of information. It need not be complete with all important
properties, birds which fly, etc., it might be a robin sitting on a branch
in sunshine,
just any bit of information, abstract, concrete, mixture of both,
whatever.
 \xEJ
\subsection{
Some (Simplified) Facts About Neurons
}

Neurons have (often many) dendrites, one core (soma), and one axon.
Dendrites have often many branches, so do axons. Basically, signals arrive
at
the dendrites, travel through the core, and are emitted through the axon.
The connection between two neurons is called a synapse, the axon of neuron
1
leads to a synapse, where it connects to a dendrite of neuron 2.
The synapse can be excitatory (positive), or inhibitory (negative).
Thus, a neuron may send a positive signal $(+1)$ to another neuron via
a positive synapse, -1 via a negative synapse, or no signal at all, 0,
when there is no connection, or the neuron does not fire. So, it is
$+1,$ -1, or 0.

Remark: A negative synapse does not correspond (directly) to negation.
When group $N_{1}$ codes ``black'', $N_{2}$ codes ``white'', then a negative
connection
between $N_{1}$ and $N_{2}$ in both ways blocks activity of the other
group,
when one is active. E.g., $N_{2}$ does not code ``not black'', as ``yellow''
is
not black either, but not coded by $N_{2}.$ Negative links rather work
for coherence.

If, within a
certain time interval, the sum of positive signals $ \xbS^{+},$ i.e. from
positive
synapses, arriving at the dendrites
of a neuron is sufficiently bigger than the sum of negative signals $
\xbS^{-},$ i.e.
from negative synapses,
arriving at the dendrites of the same neuron, the neuron will fire, i.e.
send
a signal via its axon to other neurons. This is a 0/1 reaction, it will
fire or
not, and always with the same strength. (If $ \xbS^{+}$ is much bigger
than $ \xbS^{-},$ the
neuron may fire with a higher frequency. We neglect this here.)
\subsection{
Comparison to Defeasible Inheritance Networks
}

For an overview of defeasible inheritance, see e.g.
 \cite{Sch97-2}.

Note that several neurons may act together as an amplifier for a neuron:
Say $N_{1}$ connects positively to $N_{2}$ and $N_{3},$ and $N_{2}$ and
$N_{3}$ each connect
positively to $N_{4},$ and $N_{1}$ is the only one to connect to $N_{2}$
and $N_{3},$ then
any signal from $N_{1}$ will be doubled in strength when arriving at
$N_{4},$
in comparison to a direct signal from $N_{1}$ to $N_{4}.$
 \xEh
 \xDH Consequently, direct links do not necessarily
win over indirect paths. If, in
addition, $N_{1}$ is connected negatively to $N_{4},$ then the signal from
$N_{1}$ to $N_{4}$
is 2 for positive value (indirect via $N_{2}$ and $N_{3})$ and -1 for
negative
value (direct to $N_{4}),$ so the indirect paths win.
 \xDH By the same argument, longer paths may be better.

On the other hand, a longer path has more possibilities of interference
by other signals of opposite polarity.

So length of path is no general criterion,
contrary to inheritance systems, where connections correspond to ``soft''
inclusions.
 \xDH The number of paths of the same polarity is important, in
inheritance
systems, it is only the existence.
 \xDH There is no specificity criterion, and no preclusion.
 \xDH In inheritance systems, a negative arrow may only be at the end of a
path, it cannot continue through a negative arrow. Systems of neurons
are similar: If a negative signal has any effect, it prevents the
receiving neuron from firing, so this signal path is interrupted.
 \xDH Neurons act directly sceptically - there is no branching into
different extensions.
 \xEj
\subsection{
The Elements of Human Reasoning
}

It seems that humans (and probably other animals) think in scenarios,
prototypes, pictures, etc.,
which are connected by association, reasoning, developments, etc.
We do not seem to think in propositions, models, properties, with the
help of logical operators, etc. This will be secondary, by people who
are educated in logic, children who learnt to speak a language, etc.

We have to look at above scenarios or pictures. For simplicity, we
call them all ``pictures''. Pictures can be complex, represent developments
over time, can be combined, analysed, etc., they need not be precise,
may be inconsistent, etc. We try to explain this, looking simultanously
at the thoughts, ``meanings'' of the pictures,
and the underlying neural structures and processes.

We imagine these scenarios etc. to be realised on the neural level by
neurons or groups of neurons, and the connections between pictures also by
neurons, or bundles of neurons.

To summarize, we have

 \xEh
 \xDH Pictures,

on the
 \xEI
 \xDH
meaning level, they correspond to thoughts, scenarios, pictures, etc.
 \xDH
neural level, they correspond to neurons, clusters of connected neurons,
etc.
 \xEJ
 \xDH Connections or paths,

on the
 \xEI
 \xDH
meaning level, they correspond to associations, deductions, developments,
etc.
 \xDH
neural level, they correspond to neurons, bundles of more or less
parallel neurons, connecting groups of neurons, etc.
 \xEJ
 \xEj

All that follows is relative to this assumption about human reasoning.
\subsubsection{
Pre-semantics and Semantics
}

In logic, a sentence like ``it rains'', or ``if it were to rain, $ \xfI $
would
take an umbrella'' has a semantics, which describes a corresponding state
in
the world.

We describe here what happens in our brain - according to our hypothesis -
and what corresponds to this ``brain state'' in the real world.
Thus, what we do is to describe an intermediate step between
an expression in the language, and the semantics in the world.

For this reason, we call this intermediate step a pre-semantics.

In other words,
real semantics interpret language and logic in (an abstraction of) the
world. Pre-semantics is an abstraction of (the functioning of) the brain.
Of course, the brain is ``somehow'' connected to the world, but this
would then be a semantics of (the functioning of) the brain.
Thus, this pre-semantics is an intermediate step between language and the
world.
\subsection{
Pictures: the Language and the Right Level of Abstraction
}
\subsubsection{
The Level of Meaning
}

A picture can be complex, we can see it as composed of sub-pictures.
A raven eats a piece of cheese, so there are a raven, a piece of cheese,
perhaps some other objects in this picture. The raven has a beak, etc.

It is not clear where the ``atomic'' components are. On the neural level, we
have
single neurons, but they might not have meaning any more.

To solve this, we postulate that there are no atomic pictures, we can
always decompose and analyse. For our purposes, this seems the best
way out of the dilemma.

In one context, the raven eating the cheese is the right level of
abstraction,
we might be interested in the behaviour of the raven. In a different
context, it might be the feathers, the beak of the animal, or the taste
of the cheese. Thus, there is no uniform adequate level of abstraction, it
depends on the context.
\paragraph{
Relation to Models
}

Pictures will usually not be any models in the logical sense.
There need not be any language defined, some parts may be complex and
elaborate,
some parts may be vague or uncertain, or only rough sketches, different
qualities like visual, tactile, may be combined. Pictures may also be
inconsistent.
\subsubsection{
The Neural Level
}

On the neural level, pictures will usually be realised by groups of
neurons,
consisting perhaps of several thousands neurons. Those groups will have an
internal
coherence, e.g. by strong internal positive links among their neurons.
But they will not necessarily have a ``surface'' like a cell wall
to which other cells or viruses may attach. The links between groups
go (basically) from
all neurons of group 1 to all neurons of group 2. There is no exterior vs.
interior, things are more flexible.

If group 1 ``sees'' (i.e. is positively connected to) all neurons of group
2,
then group 2 is the right level of abstraction relative to group 1 (and
its
meaning). (In our example, group 1 looks at the behaviour of the raven.)
If
group 1a ``sees'' only a subgroup of group 2 (e.g. the feathers of the
raven),
either by having particularly strong connections to this subgroup, or
by having negative connections to the rest of group 2, then
this subgroup is the right level of abstraction relative to group 1a.
Thus, the ``right'' level of abstraction is nothing mysterious, and does
not depend on our speaking about the pictures, but is given by the
activities of the neuron groups themselves.

(We neglect that changing groups of neurons may represent
the same pictures.)
\subsubsection{
The Conceptual Difficulties of this Idea
}

This description has certain conceptual difficulties.

 \xEh
 \xDH In logic, we have atoms (like propositional variables), from which
we
construct complex propositions with the use of operators like $ \xcu,$ $
\xco,$ etc.
Here, our description is ``bottomless'', we have no atoms, and can
always look inside.
 \xDH There is no unique adequate level of abstraction to think about
pictures. The right granularity depends on the context, it is
dynamic.
 \xDH The right level of abstraction on the neural level is not given by
our
thoughts about
pictures, but by the neural system itself. Other groups determine the
right granularity.
 \xDH Groups of neurons have no surface like a cell or a virus do, there
is
no surface from which connections arise.
Connections go from everywhere.
 \xEj
\subsubsection{
Summary
}

 \xEI
 \xDH
``Pictures'' on the meaning level correspond to (coherent) groups of
neurons.
 \xDH
There are no minimal or atomic pictures and groups, they can always
be decomposed (for our purposes). Single neurons might not have
any meaning any more.
 \xDH
Conversely, they can be composed to more complex pictures and groups.
 \xDH
Groups of neurons have no surface, connections to other groups are
from the interior.

 \xEJ
\subsection{
Connections or Paths
}

One neuron can connect to another neuron via axon, synapse, dendrite.
This is the simplest connection.

Connections may also be indirect, $N_{1}$ via $N_{2}$ to $N_{3},$ etc.
Groups of neurons may connect to other groups of neurons in various ways,
some direct, some indirect, etc., by single neurons and synapses, or
by bundles of neurons and synapses.

We sometimes call all such connections paths.

Connections may correspond to many different things on the meaning level.
They may be:
 \xEh
 \xDH arbitrary associations, e.g. of things which happened at the same
moment,
 \xDH inferences, classical or others,
 \xDH connections between related objects or properties, like between
people
and their ancestors, animals of the same kind, etc.,
 \xDH developments over time, etc.
 \xEj

Again, there are some conceptual problems involved.

 \xEh
 \xDH As for pictures, it seems often (but perhaps less dramatically)
difficult to
find atomic connections. If group $N_{1}$ is connected via path $P$ to
group
$N_{2},$ but $N'_{1}$ is a sub-group of $N_{1},$ connected via a subset
$P' $ of $P$ to sub-group
$N'_{2}$ of $N_{2},$ then it may be reasonable to consider $P' $ as a
proper path itself.
 \xDH If, e.g., the picture describes a development over time, with
single pictures at time $t,$ $t',$ etc. linked via paths expressing
developments,
then we have paths
inside the picture, and the picture itself may be considered a path
from beginning to end.

Thus, paths may be between pictures, or internal to pictures, and there is
no
fundamental distinction between paths and pictures.
It depends on the context.
More abstractly, the
whole path is a picture, in more detail, we have paths between single
pictures, ``frames'', as in a movie.

 \xEj

Remember: Everything is just suitably connected neurons!
\subsubsection{
Comments on Groups and Connections
}

 \xEh
 \xDH The use of a group of neurons makes this group easier accessible,
and strengthens its internal coherence.

Thus, the normal case becomes stronger.
 \xDH The use of a neural connection strengthens this connection.
Again, the normal connections become stronger.
Both properties favour learning, but may also lead to overly
rigid thinking and prejudice.

Note that this is the opposite of basic linear logic, where the use of
an argument consumes it.
 \xDH When two groups of neurons, $N_{1}$ and $N_{2}$ are activated
together,
this strengthens the connection between $N_{1}$ and $N_{2}.$ We call this
the $ \xbD $-rule. See e.g.  \cite{Pul13}.

This property establishes associations. When $ \xfI $ hear a roar in the
jungle,
and see an attacking tiger, next time, $ \xfI $ will think ``tiger'' when $
\xfI $ hear
a roar, even without seeing the tiger.

 \xDH A longer connection may be weaker. For instance, penguins are an
abnormal subclass of birds. Going from birds to penguins will not be
via a strong connection (though $penguin \xcp bird$ is a classical
inference).
If Tweety is a penguin, we might access
Tweety only by detour through penguin. Thus, Tweety is ``less''
bird than the raven which $ \xfI $ saw in my garden.
Consequently, the subset relation involved in the properties of many
nonclassical logics
has a certain ``cost'', and the resulting properties
(e.g. $X \xcc Y \xcp \xbm (Y) \xcs X \xcc \xbm (X)$ for basic preferential
logic, similar
properties for theory revision, update, counterfactual conditionals)
cannot always be expected.
 \xEj
\subsubsection{
Summary
}

 \xEI
 \xDH
Connections are made of single axon-synapse-dendrite tripels, connecting
one neuron to another, or many such tripels, bundels, or composed
bundels. Again, it seems useful to say that connections can usually be
decomposed into sub-connections.
 \xDH
Connections can be via excitatory or inhibitory synapses, the former
activate the downstream neuron, the latter de-activate the downstream
neuron.
 \xDH
Connections can have very different meanings.
 \xDH
Connections can be interior to groups, or between groups.
 \xDH
There usually is no clear distinction between groups and connections.

 \xEJ
\subsection{
Operations on Pictures and Connections
}

To simplify, we will pretend that operations are composed of
cutting and composing. We are aware that this is probably
artificial, and, more generally, an operation takes one or more
pictures (on the meaning level) or groups (on the neural level),
and constructs one or more new pictures or groups.

Before we describe our ideas, we discuss attention.
\subsubsection{
Attention
}

An additional ingredient is ``attention''.
We picture attention as a light which shines on some areas of the brain,
groups of neurons, perhaps only
on parts of those areas, and their connections, or only parts of the
connections.

Attention allows, among other things, to construct a seemingly coherent
picture
by focussing only on parts of the picture, which are coherent.
In particular, we might focus our attention on coherences, e.g., when we
want to consolidate a theory, or incoherences, when we want to attack a
theory. Focussing on coherences might hide serious flaws in a theory,
or our thinking in general. In context $ \xCf A,$ we might focus on $ \xba
,$ in context
$B,$ on $ \xbb,$ etc.

As we leave attention deliberately unregulated, changes in attention may
have very ``wild'' consequences.

Activation means that the paths leading to the picture become more active,
as well as the internal paths of the picture.

Thus, whereas memory (recent use) automatically increases activity,
attention
is an active process.

Conversely, pictures which are easily accessible (active paths
going there), are more in the focus of our attention.
E.g., we are hungry, think of a steak (associative memory), and
focus our attention on the fridge where the steak is.

Attention originates in the ``$ \xfI $'' and its aims and desires.
Likewise, ``accessibility'' is relative to the ``$ \xfI $'' - whatever
that means.
(This is probably a very simplistic picture, but suffices here.
We conjecture that the ``$ \xfI $'' is an artifact, a dynamic
construction,
with no clear definition and boundary. The ``$ \xfI $'' might be just as
elusive
as atomic pictures.)
Attention is related to our aims (find food, avoid dangers, etc.)
and allows to focus on certain pictures (or parts of pictures) and
paths.
\subsubsection{
Operations
}

Consider the picture of a raven eating a piece of cheese.

We might focus our attention on the raven, and neglect the cheese. It is
just
a raven, eating something, or not. So the connection to the raven part
will
be stronger (positive), to the cheese part weaker positive and/or stronger
negative.

Conversely, we might never have seen a raven eat a piece of cheese.
But we can imagine a raven, also a raven eating something, and a piece
of cheese, and can put these pictures together. This may be more or less
refined, adjusting the way the cheese lies on the ground, the raven
pecks at it, etc. It is not guaranteed that the picture is consistent,
and we might also adjust the picture ``on the fly'' to make it consistent
or plausible.

(When composing ``raven'' with ``cheese'' and ``pecking'', the order might be
important: Composing ``raven'' first with ``pecking'' and the result
``raven $+$ pecking'' with ``cheese'', or ``raven'' with ``pecking $+$ cheese''
might a priori give different results.)

It is easy to compose a picture of an elefant with the picture of wings,
and to imagine an elefant with small wings which hovers above the ground.
Of course, we know that this is impossible under normal circumstances.
There is no reality check in dreams, and a flying elefant is quite
plausible.

This is all quite simple (in abstract terms), and everyone has done it.
Details need to be filled in by experimental psychologists and
neuroscientists.

Obviously, these problems are related to planning.
\paragraph{
Identification vs Description
}

We may address a (part of a) picture without having a description of it,
an identification suffices.

Suppose we see an animal, and hear some strange frightening noise.
We can address the noise by the label ``that scary noise'' without being
able to describe the noise more precisely. We can isolate, compose, etc.
this
part of the picture, the identification is enough. In above scenario, we
used
descriptions of elefants, wings, etc., but this is not always necessary.
\section{
Analysis of Counterfactual Conditionals
}

We apply our ideas to counterfactuals.

Note that the Stalnaker/Lewis semantics hides all problems
in the adequate notion of distance, so we should not expect miracles
from our approach.
\subsection{
The Umbrella Scenario
}

``If it were to rain, $ \xfI $ would take an umbrella.''

We have the following present situations, where the sentence is
uttered.

 \xEh
 \xDH
Case 1: The ``normal'' case. No strong wind, $ \xfI $ have at least one hand
free to
hold an umbrella, $ \xfI $ do not want to get wet, etc.
 \xDH
Case 2: As case 1, but strong wind.
 \xDH
Case 3: As case 1, but $ \xfI $ carry things, and cannot hold an umbrella.
 \xEj

We have the following pictures in our memory:
 \xEh
 \xDH
Picture 1: Normal weather, it rains, we have our hands free, but forgot
the umbrella, and get soaked.
 \xDH
Picture 2: As picture 1, but have umbrella, stay dry.
 \xDH
Picture 3: Rain, strong wind, use umbrella, umbrella is torn.
 \xDH
Picture 4: As picture 1, but carry things, cannot hold umbrella, get
soaked.
 \xDH
Picture 5: The raven eating a piece of cheese.
 \xEj

Much background knowledge goes into our treatment of counterfactuals.
For instance, that a strong wind might destroy an umbrella (and that
the destruction of an umbrella in picture 3 is not due to some
irrelevant aspect), that we need
at least one hand free to hold an umbrella, that we want to stay dry,
that we cannot change the weather, etc.

First, we actively (using attention) look for pictures which have
something to
do with
umbrellas. Thus, in all cases, picture 5 is excluded.

Next, we look at pictures which support using an umbrella, and those which
argue against this. This seems an enormous amount of work, but our
experience tells us that a small number of scenarios usually give
the answer. There are already strong links to those scenarios.

Case 1: Pictures 3 and 4 do not apply - they are too distant in the
Stalnaker/Lewis terminology.
So we are left with pictures 1 and 2. As we want to stay dry, we choose
picture 2.
Now, we combine case 1 with picture 2 by suitable connections, and
``see'' the imagined picture where we use an umbrella and stay dry.

Case 2: Pictures 1 and 4 do not apply. $ \xfI $ would prefer to stay dry,
but
a torn umbrella does not help. In addition, $ \xfI $ do not want my
umbrella
to be torn. Combining case 2 with picture 3 shows that the umbrella
is useless, so $ \xfI $ do not take the umbrella.

Case 3: Only picture 4 fits, $ \xfI $ combine and see that $ \xfI $ will
get wet,
but there is nothing $ \xfI $ can do.
\subsection{
A Tree Felling Scenario
}

Consider the sentence:

``If $ \xfI $ were to fell that tree, $ \xfI $ would hammer a pole into the
ground,
and tie a rope between tree and pole, so the tree cannot fall on the
house.''

We have the present situation where the tree stands close to the
house, there are neither rope nor pole, nor another solid tree where we
could
anchor the rope, and we do not want to fell the tree.

We have

 \xEI
 \xDH Picture 1 of a pole being hammered into the ground - for instance,
we remember this from camping holidays.

 \xDH Picture 2 of a rope tied to a tree and its effect - for instance,
we once fastened a hammock between two young trees and saw the effect,
bending the trees over.

 \xDH Picture 3 of someone pulling with a rope on a big tree - it did not
move.
 \xEJ

We understand that we need a sufficiently strong force to prevent the
tree from falling on the house.

Pictures 2 and 3 tell us that a person pulling on the tree, or a rope
tied to a small tree will not be sufficient.

As there is no other sturdy tree around, we have to build a complex
picture. We have to cut up the hammock Picture 2 and the tent Picture 1,
using the rope part from Picture 2, the pole part from Picture 1.
It is important that the pictures are not atomic.
Note that we can first compose the situation with the pole part, and the
result with the rope part, or first the situation with rope part, and the
result
with the pole part,
or first combine the pole part with the
rope part, and then with the situation.
It is NOT guaranteed that the outcome of the different ways will be the
same.
When there are more pictures to consider, even the choice of the
pictures might depend on the sequence.
\section{
Comments
}

There are many aspects we did not treat.
We established a framework only.

 \xEh
 \xDH Usually, there are many pictures to choose. How do we make the
choice?
 \xDH How do we cut pictures?
 \xDH How do we determine if a combined picture is useful?
 \xDH Attention can hide inconsistencies, or focus on inconsistencies,
how do we decide?
 \xDH Are all these processes on one level, or is it an interplay
between different levels (execution and control)?
 \xDH These processes seem arbitrary, but we are quite successful, so
there
must be a robust procedure to find answers.
 \xEj

Some of the answers will lie in the interplay between (active) attention
and more passive memory (more recent and more frequently used pictures and
processes are easier accessible).
Recall here Edelman's insight, see e.g.
 \cite{Ede89},  \cite{Ede04},
to see the parallels between the brain and the immune
system, both working with selection from many possibilities.
We assume that we
have many candidates of the same type, so we have a population from
which to chose. We chose the best, and consider this set
for the properties of those combined areas.

It is natural to combine the ideas of the hierarchy in
 \cite{GS16}, chapter 11 there,
with our present ideas. Exceptional classes, like penguins,
are only loosely bound to regular classes, like birds; surprise
cases even more loosely.

\end{document}